\documentclass[letterpaper,journal,10pt]{IEEEtran}

\usepackage[utf8]{inputenc}
\usepackage[english]{babel}
\usepackage{booktabs}
\usepackage{amssymb, comment}
\usepackage[table,xcdraw]{xcolor}
\usepackage{cite}
\usepackage{float}
\usepackage[pdftex]{graphicx}
\DeclareGraphicsExtensions{.pdf,.png}
\usepackage{caption}
\usepackage{subcaption}
\usepackage{balance}
\usepackage[cmex10]{amsmath}
\usepackage{soul}
\usepackage{color}
\usepackage{lipsum}
\usepackage{placeins}

\usepackage{textcomp}
\usepackage{rotating}
\usepackage{url}
\usepackage{xurl}
\usepackage{hyperref}

\usepackage{booktabs}  
\usepackage{array}     

\usepackage{epstopdf}
\usepackage{graphicx}

\makeatletter
\def\section{\@startsection{section}{1}{\z@}{0.2ex plus 0.2ex minus 0.2ex}%
{0.4ex plus 0.4ex minus 0ex}{\normalfont\normalsize\centering\scshape}}%
\def\subsection{\@startsection{subsection}{2}{\z@}{0.2ex plus 0.2ex minus 0.2ex}%
{0.4ex plus .3ex minus 0ex}{\normalfont\normalsize\itshape}}%
\def\subsubsection{\@startsection{subsubsection}{3}{\parindent}{0ex plus 0.1ex minus 0.1ex}%
{0ex}{\normalfont\normalsize\itshape}}%
\def\@listi{\leftmargin\leftmargini
  \topsep 2\p@ \@plus1\p@ \@minus1\p@
  \parsep \z@
  \itemsep 1\p@ \@plus0.5\p@ \@minus0.5\p@}
\let\@listI\@listi
\makeatother

\AtBeginDocument{%
  \setlength\abovedisplayskip{3pt plus 2pt minus 2pt}%
  \setlength\belowdisplayskip{3pt plus 2pt minus 2pt}%
  \setlength\abovedisplayshortskip{1pt plus 2pt}%
  \setlength\belowdisplayshortskip{2pt plus 2pt minus 1pt}%
}

\begin{document}
\bstctlcite{IEEEexample:BSTcontrol}

\title{FI-TW: An Open Train--Weather Dataset for \\ Railway Delay Analysis in Finland
\author{Vinicius~Pozzobon~Borin, Jean~Michel~de~Souza~Sant'Ana, Usama Raheel,  and Nurul~Huda~Mahmood
\thanks{V. P. Borin, J. M. S. Sant'Ana, U. Raheel and N. H. Mahmood are with the Centre for Wireless Communications (CWC), University of Oulu, Finland (\{vinicius.borin, jean.desouzasantana, nurulhuda.mahmood\}@oulu.fi). Corresponding author: vinicius.borin@oulu.fi}
\thanks{This research was partially supported in Finland and Sweden by the European Union through the Interreg Aurora project ENSURE-6G (Grant Number: 20361812); and in Finland by the Research Council of Finland through the project 6G Flagship (Grant Number: 369116) and by the European Commission through the Horizon Europe/JU SNS project Hexa-X-II (Grant Number: 101095759)}}}

\maketitle

\begin{abstract}
Train delays result from complex interactions between operational, technical, and environmental factors. Weather strongly affects railway reliability in Nordic regions, yet publicly available datasets rarely integrate meteorological information with operational train records, which limits research on weather-driven delay. We constructed the Finland Integrated Train--Weather (FI-TW) dataset by combining operational records from the Finland Digitraffic Railway Traffic Service with observations from 209 Finnish Meteorological Institute stations, covering January 2018 to December 2024. Train events and weather measurements were aligned in space and time using the Haversine distance, with a radial fallback strategy that recovers missing parameters from alternative nearby stations. Processing included cyclical encoding of temporal features, robust scaling of weather data to limit the effect of sensor outliers, duplicate removal, and the derivation of weather-scenario indicators together with multi-scale rolling-window aggregations. The dataset contains 138 features spanning operational variables and meteorological measurements, and approximately 38.5 million observations from Finland's $5{,}915$-kilometer rail network. Exploratory analysis shows a clear seasonal structure, with winter delay rates exceeding 25\% compared with below 20\% in summer, and geographic clustering of high-delay corridors in central and northern Finland. A baseline experiment using extreme gradient boosting (XGBoost) regression reached a mean absolute error of 2.73 minutes for station-specific delay prediction. FI-TW is, to the best of our knowledge, the first publicly available dataset that couples Finnish railway operations with synchronized meteorological observations over a seven-year span. It offers multiple target variable formulations and supports applications such as delay prediction, weather impact assessment, seasonal reliability analysis, and infrastructure vulnerability mapping, providing a reusable resource for machine learning research on railway operations in severe northern climates.
\end{abstract}


\section{Background}

\IEEEPARstart{R}{ailway} transportation constitutes one of the most widely utilized modes of transportation globally, serving billions of passengers annually and facilitating the movement of goods across vast distances~\cite{Gkoumas2023}. From the high-speed rail networks of Asia to the transcontinental systems of North America, railways form the backbone of sustainable mobility in both developed and developing nations. In Europe particularly, rail transportation plays a vital role in economic infrastructure and cultural connectivity, with the continent's extensive network spanning over $245{,}000$ kilometers. This network serves millions of passengers daily, connecting major urban centers across diverse geographical and climatic conditions. 
As modern societies increasingly prioritize sustainable transportation alternatives to reduce carbon emissions and traffic congestion, railway systems have become critical infrastructure for both daily commuting and long-distance travel. Their reliability is therefore a key concern for passengers, government agencies, and private companies.

Within the scope of railway logistics, train delays represent one of the most significant concerns facing operators and passengers worldwide, caused by an interplay of technical, operational, environmental, and human factors~\cite{Mukunzi2024}. Infrastructure-related issues such as track conditions, signaling system failures, and rolling stock malfunctions can cascade through interconnected networks, affecting multiple services simultaneously. Operational factors, including schedule optimization, crew management, and station dwell times, further compound these challenges. In addition, external elements such as adverse weather conditions, passenger behavior during peak hours, and unexpected incidents introduce additional variability~\cite{Mukunzi2024}. The ability to accurately predict and analyze these delays has significant implications for both operational efficiency and passenger satisfaction. For railway operators, delay prediction enables proactive resource allocation, dynamic rescheduling, and improved maintenance planning. For passengers, reliable delay information facilitates better travel planning, reduces uncertainty, and minimizes the inconvenience associated with disrupted journeys~\cite{Monsuur2021}.

The convergence of advanced wireless communication technologies and artificial intelligence (AI) presents unprecedented opportunities for transforming railway operations. Next-generation networks such as 5G and emerging 6G systems enable real-time, high-bandwidth data collection from distributed sensors across railway infrastructure and station facilities~\cite{6gwhitepaper}. This enhanced connectivity facilitates continuous monitoring of critical parameters, including track conditions, vehicle performance metrics, and passenger flows. It thereby creates rich datasets that were previously difficult or impossible to collect centrally and analyze in real time. When combined with modern AI techniques, particularly deep learning (DL) and predictive analytics, these data streams enable railway operators to anticipate potential failures, optimize maintenance schedules, and predict delays before they occur. Machine learning (ML) models trained on comprehensive operational data can identify complex patterns indicative of emerging problems, allowing for proactive interventions rather than reactive responses. This shift from traditional scheduled maintenance to predictive, condition-based approaches not only reduces operational costs but also significantly improves service reliability and safety across the entire railway network~\cite{Davari2021}.

The availability of train-related datasets has expanded in recent years across four primary thematic categories: traffic planning and management, maintenance and inspection, safety and security, and passenger mobility~\cite{Pappaterra2021}. However, a significant limitation persists: most existing datasets either omit weather conditions entirely or incorporate only a limited set of meteorological parameters. This gap presents a critical challenge for research requiring simultaneous analysis of operational and meteorological factors. Such analysis is essential given the potential impact of weather on railway performance.

\subsection{Related Work in Railway Research} \label{subsec:sota_related_work}

A comprehensive systematic review~\cite{Pappaterra2021} examined 62 publicly available AI-oriented datasets for railway applications, finding that while operational data dominates the field (47 of 62 datasets), the vast majority do not incorporate meteorological data. Among the few that couple railway data with environmental conditions, RailSem19~\cite{Zendel2019} provides $8{,}500$ annotated sequences for semantic scene understanding, complementing the Cityscapes dataset~\cite{cordts2016cityscapes_web}, while the Indian Metro Dataset~\cite{ansari2019indian} pairs traffic and passenger-flow data with a limited set of weather conditions (humidity, wind, visibility, and precipitation). Both, however, capture only a narrow range of meteorological parameters.

More recently, a small number of datasets have begun to integrate operational records with weather observations more explicitly. Zhang et al.~\cite{Zhang2022} released a Chinese high-speed railway dataset covering 727 stations and $3{,}399$ trains over a 16-week window (2019--2020), annotated with weather condition, temperature, and wind level. Wu et al.~\cite{Wu2026} published an Italian railway network dataset spanning $1{,}397$ stations and $3{,}324$ trains over six months of 2024, combining multi-type train operation records with temperature, wind, and general weather conditions retrieved from a public weather service and, as in our work, using the Haversine formula to compute inter-station distances. At nationwide scale, the RIDE benchmark~\cite{Elliker2026} links $94.5$~million Belgian train events with $35.7$~million weather records (2023--2025) and provides a standardized evaluation protocol for delay-prediction models. These efforts confirm the growing recognition that weather is integral to railway analytics; nevertheless, they remain constrained by comparatively short temporal spans, a narrow set of meteorological variables, and none captures the sub-Arctic conditions that dominate Nordic rail operations. The dataset proposed here is distinguished by its seven-year span (2018--2024), its breadth of meteorological coverage (13 weather parameters underlying 138 features, including engineered weather-scenario indicators and multi-scale rolling-window aggregations), its fine-grained station-level spatial matching with a radial fallback strategy, and its focus on Finland's severe northern climate.

Recent delay prediction studies have begun incorporating weather factors. Huang et al.~\cite{Huang2020} developed FCLL-Net for Chinese high-speed railways, achieving 9.4\% improvement by capturing train interactions alongside temperature, wind, and rainfall. Another study~\cite{Huang2021} applied cost-sensitive deep learning (FCF-Net) to model timetables as images for delay propagation pattern recognition. Sajan et al.~\cite{Sajan2021} evaluated regression models for Indian Railways, finding elastic-net regression effective when calibrated with weather variables, though departure delay remained the strongest predictor. Notably, Oneto et al.~\cite{oneto2018train} achieved significant improvements using Deep Extreme Learning Machines on Italian railway data, but explicitly identified weather data integration as critical future work. In Nordic contexts, where winter weather is especially disruptive, Zakeri and Olsson~\cite{Zakeri2018} analysed ten years of punctuality data from Norway's Nordland Line and identified snow depth as the strongest weather-related predictor of low punctuality, with extreme cold winters driving the largest delays. More recently, Soleimani-Chamkhorami et al.~\cite{Soleimani2024} fused Swedish infrastructure-failure records with meteorological observations to classify climate-related failures through machine learning, reporting snow, ice, and low temperature as the dominant factors. Collectively, these studies underscore the value of a richly featured, weather-integrated dataset for Nordic railway research, yet each relies on proprietary or task-specific data rather than a reusable public resource.

\subsection{Proposed Dataset} \label{subsec:contribution}

To address the identified gap in integrated train-weather datasets, this work introduces the Finland Integrated Train--Weather (FI-TW) dataset, a novel resource from the Finnish railway system that systematically combines operational train data with comprehensive meteorological information. The Finnish railway network comprises over $5{,}900$ kilometers of track and serves over 90 million passengers annually~\cite{ftia2024railway}. It experiences diverse climatic conditions ranging from Nordic coastal regions to severe Arctic weather, making it an interesting context for studying weather impacts on railway operations.

The dataset integrates temporal operational metrics, like train schedules, train types, routes, and delays, with synchronized weather observations, enabling comprehensive analysis of environmental impacts on railway performance. The main contributions of this work are:

\begin{itemize}
    \item The development of the first (to the best of our knowledge) publicly available dataset integrating operational data with meteorological data for the Finnish railway network.
    \item Step-by-step data processing to create a cohesive dataset for ML applications.
    \item A comprehensive spatial-temporal coverage across diverse climatic conditions.
    \item A multi-dimensional integration of operational and meteorological variables with precise temporal synchronization.
    \item The consideration of multiple research applications beyond delay prediction: weather impact assessment, seasonal reliability analysis, weather-adaptive scheduling, and infrastructure vulnerability mapping.
\end{itemize}

\section{Collection Methods and Design} \label{sec:method}

This study focuses on long-distance services, which connect major cities across Finland over main lines spanning several hundred kilometers. According to VR Group, the country's primary passenger rail operator, more than 15 million of Finland's passenger journeys in 2024 were long-distance~\cite{vr_longdistance_2025}. A representative route is the 875-kilometer line that traverses the country, connecting the capital Helsinki in the south to the arctic city Rovaniemi in the north.

Punctuality for these services is assessed by the Finnish Transport Infrastructure Agency based on arrival at the final destination, where a train is considered on time if it arrives within five minutes of its scheduled time. In 2024, 86.28\% of long-distance trains met this standard~\cite{vayla2025railway}. Maintaining such punctuality across an extensive network is challenging, as signal and track equipment malfunctions, including power failures, communication breakdowns, and point failures, frequently disrupt operations until repairs are completed.

Finland's extreme climate poses particularly severe challenges, as winter temperatures reaching $-40$°C can cause mechanical failures in automatic doors, couplings, and switching systems. Heavy snowfall disrupts signaling equipment and requires extensive track clearing operations~\cite{Lotfi2023}. During autumn, fallen leaves create slippery layers on rails, reducing adhesion and requiring trains to operate at lower speeds for safety~\cite{Lewis2023}. These examples of weather-related and technical issues often cascade through the interconnected rail network, amplifying delays across multiple routes.

To analyze these operational challenges and delay patterns systematically, this study constructs a comprehensive dataset by integrating two Finnish open data sources that are described in this section. The dataset encompasses observations collected from January 2018 to December 2024. All example data presented in this section utilize observations from a long-distance train service operating between Oulu and Helsinki on December 1\textsuperscript{st}, 2024.

\subsection{Digitraffic Railway Traffic Data}

Operated by Fintraffic, Digitraffic Railway Traffic Service\footnote{Service webpage: https://www.digitraffic.fi/en/railway-traffic/} provides comprehensive real-time and historical data about trains operating throughout the Finnish railway network. The dataset includes train timetables (scheduled and actual departure/arrival times), real-time location data (Global Positioning System [GPS] coordinates) and speed, train composition details, and station infrastructure metadata for Finland's state-owned railway network. For historical data, access is provided through an Application Programming Interface (API). The service requires no authentication and enforces a rate limit of 60 requests per minute per Internet Protocol (IP) address, with most responses cached for approximately one minute. All data are provided under the Creative Commons Attribution 4.0 International license (CC BY 4.0), enabling free use with appropriate attribution.

We retrieved historical data from the API endpoint\footnote{API Endpoint: /api/v1/trains/\{departure\_date\}}, which returns all trains operated on a given departure date. Table~\ref{tab:column_descriptions} presents the descriptors of the data retrieved from this endpoint. As shown in this table, the dataset includes a column labeled \textit{timeTableRows}, which contains an array of schedule information for each train number operating on that specific date. The structure of the timetable data is detailed in Table~\ref{tab:timetablerows_descriptions}.

\begin{table}[!htbp]
\vspace{-0.4cm}
\centering
\small
\caption{Fintraffic Dataset Descriptors}
\label{tab:column_descriptions}
\begin{tabular}{ll}
\toprule
\textbf{Column Name} & \textbf{Description} \\
\midrule
trainNumber & Unique train identification number \\
departureDate & Date of train departure \\
operatorUICCode & UIC code of train operator \\
operatorShortCode & Short code for operator (e.g., VR) \\
trainType & Type of train (e.g., IC) \\
trainCategory & Train category (e.g., Long-distance) \\
commuterLineID & Commuter line identifier \\
runningCurrently & Boolean if train is running \\
cancelled & Boolean if train is cancelled \\
version & Version number of timetable entry \\
timetableType & Type of timetable (e.g., REGULAR) \\
timetableAcceptanceDate & Date timetable was accepted \\
timeTableRows & Nested data object of station timing data \\
\bottomrule
\end{tabular}
\vspace{-0.2cm}
\end{table}

\begin{table}[!htbp]
\vspace{-0.4cm}
\centering
\small
\caption{timeTableRows Data Descriptors}
\label{tab:timetablerows_descriptions}
\begin{tabular}{ll}
\toprule
\textbf{Column Name} & \textbf{Description} \\
\midrule
stationName & Name of the railway station \\
stationShortCode & Short code identifier for the station \\
type & Event type (ARRIVAL or DEPARTURE) \\
scheduledTime & Scheduled arrival/departure time (UTC+2)\\
actualTime & Actual arrival/departure time (UTC+2)\\
differenceInMinutes & Delay in minutes (actual - scheduled) \\
cancelled & Boolean indicating if stop was cancelled \\
stationUICCode & UIC code for the station \\
countryCode & ISO country code of station location \\
trainStopping & Boolean indicating if train stops at station \\
commercialStop & Boolean indicating if stop is commercial \\
commercialTrack & Track number for commercial operations \\
causes & Array of delay/disruption causes \\
trainReady & Train readiness information \\
liveEstimateTime & Real-time estimated arrival/departure time \\
estimateSource & Source of time estimate data \\
\bottomrule
\end{tabular}
\vspace{-0.2cm}
\end{table}

Table~\ref{tab:train_schedule} presents a detailed example of train schedule data, showing only relevant columns. Each row in the table represents a track section traversed by the train, which does not necessarily correspond to a passenger station. The table includes two temporal columns: \textit{scheduledTime} and \textit{actualTime}. The \textit{differenceInMinutes} column is pre-calculated by Fintraffic and retrieved directly from the API as
\begin{equation}
\textit{differenceInMinutes} = \textit{actualTime} - \textit{scheduledTime}.
\end{equation}
This value represents the delay or early arrival at each point
\begin{equation}
\begin{cases}
\text{On time,} & \text{if } \textit{differenceInMinutes} = 0, \\
\text{Delayed,} & \text{if } \textit{differenceInMinutes} > 0, \\
\text{Ahead of schedule,} & \text{if } \textit{differenceInMinutes} < 0.
\end{cases}
\end{equation}

\begin{table}[!htbp]
\vspace{-0.4cm}
\centering
\caption{Train Schedule Data: Oulu to Helsinki}
\label{tab:train_schedule}
\resizebox{\columnwidth}{!}{%
\footnotesize
\setlength{\tabcolsep}{3pt}
\begin{tabular}{@{}rllcccrcc@{}}
\toprule
\rotatebox{90}{\#} & \rotatebox{90}{stationName} & \rotatebox{90}{stationShortCode} & \rotatebox{90}{type} & \rotatebox{90}{scheduledTime} & \rotatebox{90}{actualTime} & \rotatebox{90}{differenceInMinutes} & \rotatebox{90}{trainStopping} & \rotatebox{90}{commercialStop} \\
\midrule
1 & Oulu asema & OL & DEP & 05:49 & 05:50 & 1 & Yes & Yes \\
2 & Oulunlahti & OLL & ARR & 05:53 & 05:57 & 4 & Yes & Yes \\
3 & Oulunlahti & OLL & DEP & 05:53 & 05:57 & 4 & Yes & Yes \\
4 & Kempele & KML & ARR & 05:57 & 06:01 & 4 & Yes & Yes \\
5 & Kempele & KML & DEP & 05:57 & 06:01 & 4 & Yes & Yes \\
\vdots & \vdots & \vdots & \vdots & \vdots & \vdots & \vdots & \vdots & \vdots \\
198 & Käpylä & KÄP & ARR & 11:26 & 11:29 & 3 & Yes & Yes \\
199 & Käpylä & KÄP & DEP & 11:26 & 11:29 & 3 & Yes & Yes \\
200 & Pasila asema & PSL & ARR & 11:28 & 11:31 & 3 & Yes & Yes \\
201 & Pasila asema & PSL & DEP & 11:29 & 11:33 & 4 & Yes & Yes \\
202 & Helsinki asema & HKI & ARR & 11:35 & 11:37 & 2 & Yes & Yes \\
\bottomrule
\end{tabular}
}
\vspace{-0.2cm}
\end{table}

The \textit{type} column in Table~\ref{tab:train_schedule} indicates whether a row corresponds to an arrival (ARR) or departure (DEP) event at a given station. For intermediate stations, each stop generates two records: one for arrival and one for departure. This distinction allows for precise tracking of dwell time at stations and delay propagation throughout the journey. The origin station (Oulu asema) contains only a DEP record, while the final destination (Helsinki asema) contains only an ARR record. Consequently, a journey with $n$ stopping points generates $2n$ records for intermediate stations, plus the origin departure and destination arrival, totaling $2n-1$ rows when all stations involve commercial stops.



\subsubsection{Train Data Overview}

The geographical distribution of Finnish railway stations shows 549 stations located within Finnish territory during the data collection period (2018-2024), comprising 340 (61.9\%) non-passenger stations designated for freight/cargo handling, docks, or technical service points and 209 (38.1\%) passenger stations. 

The functional classification of stations within the Finnish network reveals a hierarchical structure with conventional stations (passengers and non-passengers) constituting 81.07\% (454 locations), stopping points representing 11.43\% (64 locations), and turnouts in the open line accounting for 7.50\% (42 locations). This distribution is characteristic of Finland's railway system, where approximately $5{,}200$ km of the $5{,}915$ km network consists of single-track lines that require strategic passing locations and auxiliary facilities for operational efficiency.

\subsubsection{Train Delays Overview}

\begin{figure}
    \centering
    \includegraphics[width=\columnwidth]{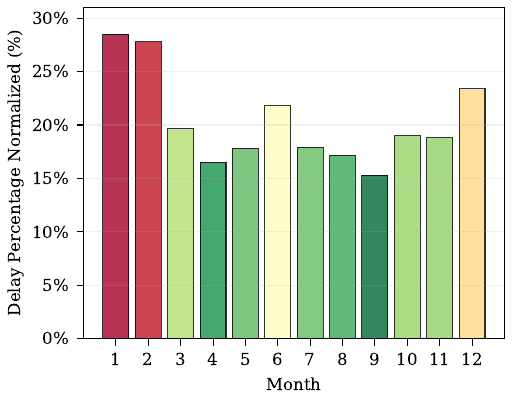}
    \caption{Aggregated Normalized Delays by month (2018-2024).}
    \label{fig:delays_month}
    \vspace{-0.8cm}
\end{figure}

The temporal analysis of delays during the period 2018-2024 in Finland reveals a distinct seasonal pattern. Figure~\ref{fig:delays_month} shows aggregated normalized delay data by month. Normalization represents the ratio of the total delays in the train schedule to the total number of train schedules. A train is considered delayed if it arrives five minutes or more after the scheduled time. Importantly, our methodology differs from the Finnish Transport Infrastructure Agency's approach: we examine each individual train schedule stop along the route rather than evaluating punctuality at the final destination only, providing a more comprehensive assessment of service reliability throughout the entire journey. It is important to note that, although the dataset flags whether each delay's officially recorded cause is weather-related (via the \textit{causes\_related\_to\_weather} feature introduced in Section~\ref{subsec:final_dataset}), the detailed underlying causes are not further resolved in this aggregated view and could be attributed to various factors such as weather-related conditions or passenger-induced delays at stations. In particular, extreme winter months (Dec-Jan-Feb) experience substantially higher delay percentages. In addition, June exhibits a relatively higher delay percentage (21.8\%) compared to other summer months, which could be related to the increased number of trains in circulation during the summer season and the higher volume of passengers traveling for holidays during this period.

\begin{figure}[!htbp]
    \centering
\includegraphics[width=\columnwidth]{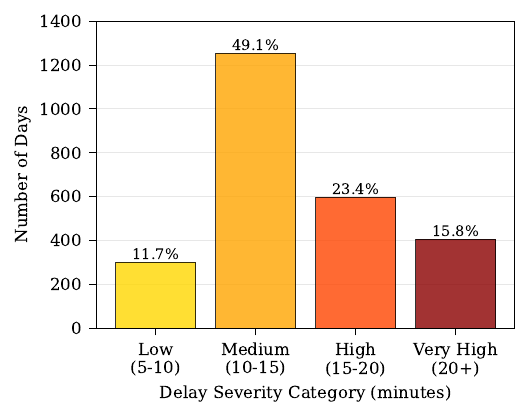}
    \caption{Distribution of delay events by delay-intensity category (2018-2024).}
    \label{fig:min_delay}
    \vspace{-0.6cm}
\end{figure}

Taking into account a range of seven years of data (2018-2024), the distribution of delay intensity categories between days is illustrated in Figure~\ref{fig:min_delay}. It reveals that medium delays (10-15 minutes) are the most prevalent,  with 1254 (49\%) occurrences, followed by high delays (15-20 minutes), with 597 (23.35\%) cases, and very high delays (more than 20 minutes) with 405 (15.84\%) occurrences. Low delays (5-10 minutes) represent the smallest category with 300 (11.73\%) entries.

\begin{figure}[!htbp]
    \centering
\includegraphics[width=\columnwidth]{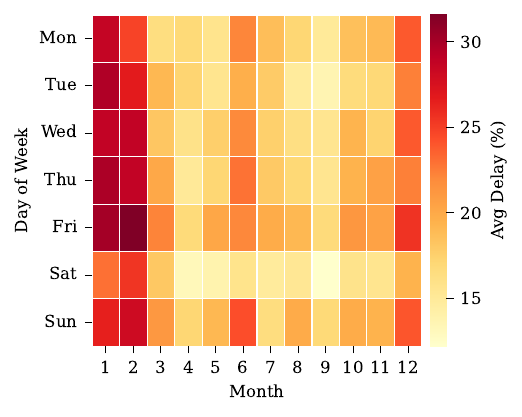}
    \caption{Average delays by week (2018 - 2024).}
    \label{fig:week_delay}
    \vspace{-16pt}
\end{figure}

The average delay percentage distribution presented in Figure~\ref{fig:week_delay} reveals weekly and seasonal patterns. Weekdays consistently exhibit higher delays compared to weekends, and Fridays demonstrate particularly high delays throughout the year, reaching up to 31.6\% in February. In contrast, Saturday demonstrates the lowest delay percentages throughout the year, averaging between 12.2\% and 25.5\%. A clear seasonal trend is again observed, with the cold months (January, February, and December) experiencing substantially higher delays throughout all weekdays, often exceeding 25\%, while the late Spring and early Autumn months (April, May, August, and September) show relatively lower delays, frequently below 20\%.

\subsection{Finnish Meteorological Institute Data} \label{subsec:fmi}

The Finnish Meteorological Institute (FMI) is a government agency under Finland's Ministry of Transport and Communications. This service offers comprehensive weather, sea, and climate observation data and forecasts alongside numerical model outputs from meteorological stations throughout Finland. 
The dataset\footnote{Service webpage: https://en.ilmatieteenlaitos.fi/open-data} encompasses comprehensive meteorological and environmental observations. It includes real-time weather measurements and weather radar data from 10 C-band Doppler radars, which provide reflectivity and precipitation intensity measurements. Marine observations cover sea level, wave height, and ocean currents. The dataset also contains air quality measurements and numerical weather prediction model outputs from the European Centre for Medium-Range Weather Forecasts (ECMWF).

Data access is provided through a Web Feature Service (WFS) API that follows international Open Geospatial Consortium standards\footnote{Query: \texttt{fmi::observations::weather::multipointcoverage}}, which provides access to multipoint weather station data. The service requires no authentication, but implements rate limits of $20{,}000$ requests per day and 600 requests per five minutes to ensure service stability. All observations are produced according to the standards of the World Meteorological Organization (WMO) with established quality control protocols. The data are provided under the GNU General Public License version 3 (GPL-3.0), allowing for free use, modification, and distribution. Table~\ref{tab:weather_columns} shows the descriptors of the weather data fetched.

\begin{table}[!htbp]
\vspace{-0.4cm}
\centering
\small
\caption{Weather Data Column Descriptions}
\label{tab:weather_columns}
\begin{tabular}{@{}ll@{}}
\toprule
\textbf{Column Name} & \textbf{Unit} \\
\midrule
timestamp & ISO 8601 \\
station\_name & string \\
Air temperature & °C \\
Wind speed & m/s \\
Gust speed & m/s \\
Wind direction & ° (degrees) \\
Relative humidity & \% \\
Dew-point temperature & °C \\
Precipitation amount & mm \\
Precipitation intensity & mm/h \\
Snow depth & cm \\
Pressure (mean sea level, msl) & hPa \\
Horizontal visibility & m \\
Cloud amount & oktas (0-9) \\
Present weather (auto) & WMO code \\
\bottomrule
\end{tabular}
\vspace{-0.2cm}
\end{table}

The \textit{Present weather (auto)} field employs the WMO Code Table 4677~\cite{wmo4677}, which encodes current atmospheric conditions as integer values ranging from 0 to 99. The code structure follows a hierarchical organization: values 00--19 indicate no precipitation with various visibility conditions (haze, mist, dust); 20--29 represent recent precipitation events; 30--39 encode dust storms, sandstorms, or blowing snow; 40--49 indicate fog conditions; and 50--99 describe active precipitation at the observation time, including drizzle (50--59), rain (60--69), snow (70--79), and showery or thunderstorm-related precipitation (80--99). This standardized encoding enables a consistent machine-readable representation of complex weather phenomena across international meteorological networks.

Table~\ref{tab:weather_data} details sample data from an Environmental Meteorological Station (EMS) located at Oulu Vihreäsaari satama\footnote{Station Coordinates. Lat: 65.006370, Long: 25.393250} on December 1\textsuperscript{st}, 2024. The Table presents all 13 weather measurements as columns. However, data for precipitation amount, precipitation intensity, and snow depth are unavailable. This absence is expected and occurs because not all EMS units measure the complete set of weather parameters. Another important detail about the data concerns the measurement time intervals, as this particular EMS records weather conditions every 10 minutes, while some stations in Finland operate at 1-minute intervals. 

\begin{table}[!htbp]
\vspace{-0.4cm}
\centering
\caption{Weather observations: Oulu Vihreäsaari satama EMS}
\label{tab:weather_data}
\resizebox{\columnwidth}{!}{%
\footnotesize
\setlength{\tabcolsep}{3pt}
\begin{tabular}{@{}lcccccccccccccc@{}}
\toprule
\rotatebox{90}{Time (UTC+2)} & \rotatebox{90}{Air Temp. (°C)} & \rotatebox{90}{Wind Speed (m/s)} & \rotatebox{90}{Gust Speed (m/s)} & \rotatebox{90}{Wind Dir. (°)} & \rotatebox{90}{Rel. Humid. (\%)} & \rotatebox{90}{Dew-point Temp. (°C)} & \rotatebox{90}{Precip. Amount (mm)} & \rotatebox{90}{Precip. Intens. (mm/h)} & \rotatebox{90}{Snow Depth (cm)} & \rotatebox{90}{Pressure (msl) (hPa)} & \rotatebox{90}{Horiz. Visib. (m)} & \rotatebox{90}{Cloud Amount (oktas)} & \rotatebox{90}{Present Weather (code)} \\
\midrule
00:00 & $-1.4$ & 1.1 & 1.4 & 28 & 97 & $-1.8$ & -- & -- & -- & 1007.1 & 10943 & -- & 71 \\
00:10 & $-1.3$ & 0.8 & 1.6 & 14 & 96 & $-1.7$ & -- & -- & -- & 1007.3 & 17765 & -- & 67 \\
00:20 & $-1.2$ & 0.8 & 1.2 & 10 & 96 & $-1.7$ & -- & -- & -- & 1007.4 & 20000 & -- & 24 \\
00:30 & $-1.2$ & 0.3 & 0.7 & 325 & 96 & $-1.8$ & -- & -- & -- & 1007.4 & 20000 & -- & 24 \\
00:40 & $-1.2$ & 0.3 & 0.6 & 9 & 96 & $-1.7$ & -- & -- & -- & 1007.4 & 20000 & -- & 24 \\
\vdots & \vdots & \vdots & \vdots & \vdots & \vdots & \vdots & \vdots & \vdots & \vdots & \vdots & \vdots & \vdots & \vdots \\
10:09 & 0.6 & 4.4 & 5.9 & 153 & 99 & 0.4 & -- & -- & -- & 1006.4 & 1867 & -- & 10 \\
10:11 & 0.6 & 4.4 & 5.9 & 153 & 100 & 0.6 & -- & -- & -- & 1006.4 & 1706 & -- & 10 \\
\vdots & \vdots & \vdots & \vdots & \vdots & \vdots & \vdots & \vdots & \vdots & \vdots & \vdots & \vdots & \vdots & \vdots \\
23:57 & 6.3 & 12.0 & 15.7 & 221 & 83 & 3.6 & -- & -- & -- & 995.8 & 20000 & -- & 0 \\
23:58 & 7.0 & 12.0 & 15.7 & 221 & 85 & 4.7 & -- & -- & -- & 995.8 & 20000 & -- & 0 \\
23:59 & 6.7 & 12.1 & 15.7 & 221 & 82 & 3.8 & -- & -- & -- & 995.8 & 20000 & -- & 0 \\
\bottomrule
\end{tabular}
}
\end{table}

\subsubsection{Weather Data Overview}

This study utilizes data from 209 operational weather stations across Finland during 2018-2024. These stations are configured to collect meteorological observations at two distinct measurement intervals: high-frequency 1-minute resolution and standard 10-minute resolution measurements. The distribution reveals that 164 stations (78.47\%) operate with 10-minute measurement intervals, representing the predominant temporal resolution in the FMI's observation network, while 45 stations (21.53\%) provide high-resolution 1-minute measurements. This temporal configuration reflects the strategic deployment of weather monitoring infrastructure, where the majority of stations employ the World Meteorological Organization's standard 10-minute averaging period for synoptic observations~\cite{WMO-No82008}, whereas high-frequency 1-minute stations are typically positioned at locations requiring detailed temporal resolution for specialized applications such as aviation meteorology~\cite{SKYbrary2024WeatherObservations}.

The instrumental capabilities and measurement coverage of the 209 environmental meteorological stations, as detailed in Table~\ref{tab:weather_coverage}, reveal a stratified distribution of meteorological parameters. Near-universal coverage (98.60\%) is observed for fundamental thermodynamic variables such as air temperature, relative humidity, and dew-point temperature, representing measurements essential for weather monitoring. Wind-related parameters demonstrate moderately high coverage at 79.40\%, while atmospheric pressure measurements exhibit 73.20\% coverage. In contrast, hydrometeorological parameters show substantially lower coverage: precipitation measurements (56.90\%), snow depth (53.60\%), cloud amount (53.10\%), and horizontal visibility (52.60\%) are recorded at approximately half of all stations. This uneven coverage across meteorological parameters substantially increases the complexity of data analysis and preparation, necessitating systematic handling strategies for missing values in subsequent preprocessing steps.

\begin{table}[!hbtp]
\vspace{-0.4cm}
\centering
\small
\caption{Weather feature measurement coverage}
\label{tab:weather_coverage}
\begin{tabular}{lc}
\toprule
\textbf{Weather Feature} & \textbf{Stations Measuring} \\
 & \textbf{(Coverage \%)} \\
\midrule
Air temperature & 206 (98.60\%) \\
Relative humidity & 206 (98.60\%) \\
Dew-point temperature & 206 (98.60\%) \\
Wind speed & 166 (79.40\%) \\
Gust speed & 166 (79.40\%) \\
Wind direction & 166 (79.40\%) \\
Pressure (msl) & 153 (73.20\%) \\
Precipitation amount & 119 (56.90\%) \\
Precipitation intensity & 119 (56.90\%) \\
Snow depth & 112 (53.60\%) \\
Cloud amount & 111 (53.10\%) \\
Horizontal visibility & 110 (52.60\%) \\
EMS with no measurements & 003 (01.43\%) \\
\midrule
\textbf{TOTAL} & \textbf{209 (100\%)} \\
\bottomrule
\end{tabular}
\vspace{-0.2cm}
\end{table}

\subsubsection{Derived Weather Features} \label{subsubsec:derived_weather}

Beyond the instantaneous observations described above, two families of derived weather features were engineered directly on the meteorological dataset, prior to its spatial--temporal merging with the train records (Section~\ref{sec:system}): \textit{rolling-window aggregations} and \textit{weather scenario indicators}. Computing these features at the weather-station level, before merging, ensures that each one summarizes the genuine local weather history recorded at the measuring station, independent of the subsequent train-matching process.

The rolling-window aggregations, summarized in Table~\ref{tab:rolling_windows}, capture the recent statistical behavior of the weather. Seven variables (air temperature, wind speed, relative humidity, precipitation intensity, snow depth, pressure, and horizontal visibility) are aggregated as minimum, maximum, and mean over 12\,h, 24\,h, and 72\,h windows, while precipitation amount is aggregated as mean and cumulative total over the same windows, yielding 69 additional columns.

\begin{table}[!hbtp]
\vspace{-0.2cm}
\centering
\small
\caption{Rolling-window aggregated weather features. Statistics are computed over 12\,h, 24\,h, and 72\,h look-back windows. Gust speed, wind direction, and dew-point temperature receive no rolling aggregation.}
\label{tab:rolling_windows}
\begin{tabular}{@{}llcc@{}}
\toprule
\textbf{Base variable} & \textbf{Statistics} & \textbf{Windows} & \textbf{Cols} \\
\midrule
Air temperature          & min, max, mean   & 12/24/72\,h & 9 \\
Wind speed               & min, max, mean   & 12/24/72\,h & 9 \\
Relative humidity        & min, max, mean   & 12/24/72\,h & 9 \\
Precipitation intensity  & min, max, mean   & 12/24/72\,h & 9 \\
Snow depth               & min, max, mean   & 12/24/72\,h & 9 \\
Pressure (msl)           & min, max, mean   & 12/24/72\,h & 9 \\
Horizontal visibility    & min, max, mean   & 12/24/72\,h & 9 \\
Precipitation amount     & mean, cumulative & 12/24/72\,h & 6 \\
\midrule
\textbf{TOTAL}           &                  &             & \textbf{69} \\
\bottomrule
\end{tabular}
\vspace{-0.2cm}
\end{table}

The weather scenario indicators, detailed in Table~\ref{tab:weather_scenarios}, encode ten named adverse-weather conditions (e.g., \textit{Blizzard}, \textit{Freezing\_Rain}, \textit{Black\_Ice}) as binary flags derived through threshold rules over the measured meteorological variables. Each scenario is evaluated at four temporal resolutions: instantaneously and over rolling 12\,h, 24\,h, and 72\,h look-back windows, where a window-level flag is set if the scenario occurs at least once within that window. This yields $10 \times 4 = 40$ columns. Together, these engineered families allow models to distinguish, for instance, a brief cold snap from a sustained sub-zero spell, a distinction that instantaneous readings alone cannot express.

\begin{table}[!hbtp]
\vspace{-0.2cm}
\centering
\small
\caption{Weather scenario indicator features. Each scenario is provided as a binary flag at four temporal resolutions: instantaneous and rolling 12\,h/24\,h/72\,h windows, with the flag set if the scenario occurs at least once within the window. This yields $10 \times 4 = 40$ columns.}
\label{tab:weather_scenarios}
\begin{tabular}{@{}ll@{}}
\toprule
\textbf{Scenario} & \textbf{Adverse-weather condition flagged} \\
\midrule
Normal\_Clear   & Baseline benign conditions, no adverse weather \\
Blizzard        & Heavy snowfall with strong winds and low visibility \\
Heavy\_Snow     & Intense snowfall / rapid snow accumulation \\
Extreme\_Cold   & Severely low air temperature \\
Heavy\_Rain     & High-intensity rainfall \\
Freezing\_Rain  & Rain onto sub-zero surfaces causing icing \\
Black\_Ice      & Near-freezing conditions favouring invisible rail ice \\
Dense\_Fog      & Strongly reduced horizontal visibility \\
High\_Winds     & Strong sustained winds and gusts \\
Extreme\_Heat   & Unusually high air temperature \\
\bottomrule
\end{tabular}
\vspace{-0.2cm}
\end{table}


\section{Validation and Quality}\label{sec:system}

Ensuring data integrity is essential when combining heterogeneous data sources, as inconsistencies in temporal alignment, spatial matching, or missing values can propagate errors through subsequent analyses and compromise model performance. The proposed dataset integrates FMI meteorological data with Digitraffic railway traffic data, specifically designed and preprocessed to facilitate machine learning applications for railway operations predictions, particularly train delay forecasting. 

\begin{figure}[!hbtp]
    \centering
    \includegraphics[width=0.7\columnwidth]{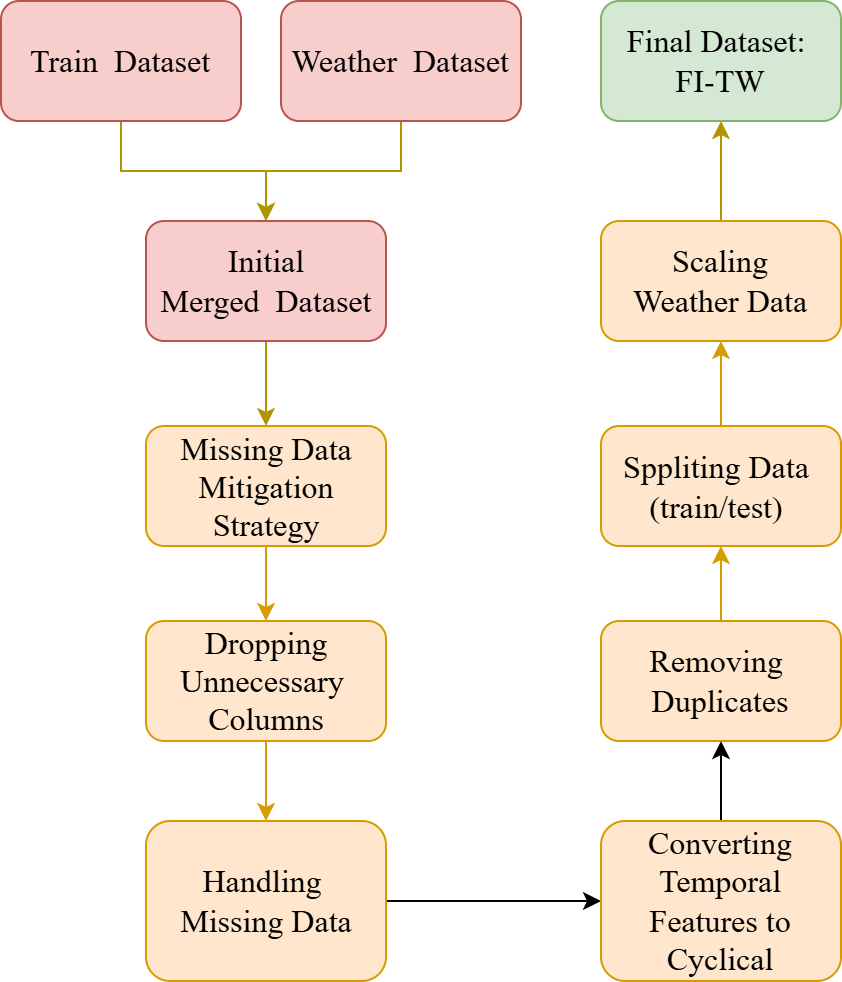}
    \caption{Flowchart of the processing steps.}
    \label{fig:dataset_creation}
    \vspace{-16pt}
\end{figure}

To the best of our knowledge, this represents the first comprehensive dataset that systematically combines train schedules with corresponding real-time weather conditions along railway routes in Finland. The methodology for constructing this integrated dataset is illustrated in Figure~\ref{fig:dataset_creation}, which outlines the sequential steps from data acquisition to final dataset assembly.

\subsection{Merging Dataset}
The integration of weather and train datasets involves a two-stage process combining spatial and temporal alignment:

\begin{enumerate}
    \item \textit{Spatial matching:} Each train section was associated with its corresponding EMS by identifying the geographically closest station using GPS coordinates. The distance between each train station and candidate EMS was computed using the Haversine formula~\cite{sinnott1984virtues}, which accounts for the spherical geometry of Earth:
    \begin{equation}
    \begin{split}
        d = 2r \arcsin\Bigg(\Big[\sin^2\Big(\frac{\phi_2 - \phi_1}{2}\Big) + \\
        \cos(\phi_1)\cos(\phi_2)\sin^2\Big(\frac{\lambda_2 - \lambda_1}{2}\Big)\Big]^{1/2}\Bigg),
    \end{split}
    \end{equation}
    where $d$ is the great-circle distance between two points, $r$ is Earth's radius (approximately $6{,}371$ km), $\phi_1$ and $\phi_2$ are the latitudes, and $\lambda_1$ and $\lambda_2$ are the longitudes of the two points in radians. The GPS coordinates metadata for train stations and EMS were obtained from their respective open databases.
    
    \item \textit{Temporal alignment:} A left join operation was performed with the train dataset as a base. For each train record, meteorological features were retrieved from its assigned EMS at the corresponding timestamp, preserving all train dataset records. For records lacking exact temporal correspondence, nearest-neighbor temporal matching was applied within a predefined tolerance window to ensure temporal relevance while maintaining data integrity.
\end{enumerate}

\begin{figure}[!htbp]
    \centering
    \includegraphics[width=0.98\columnwidth]{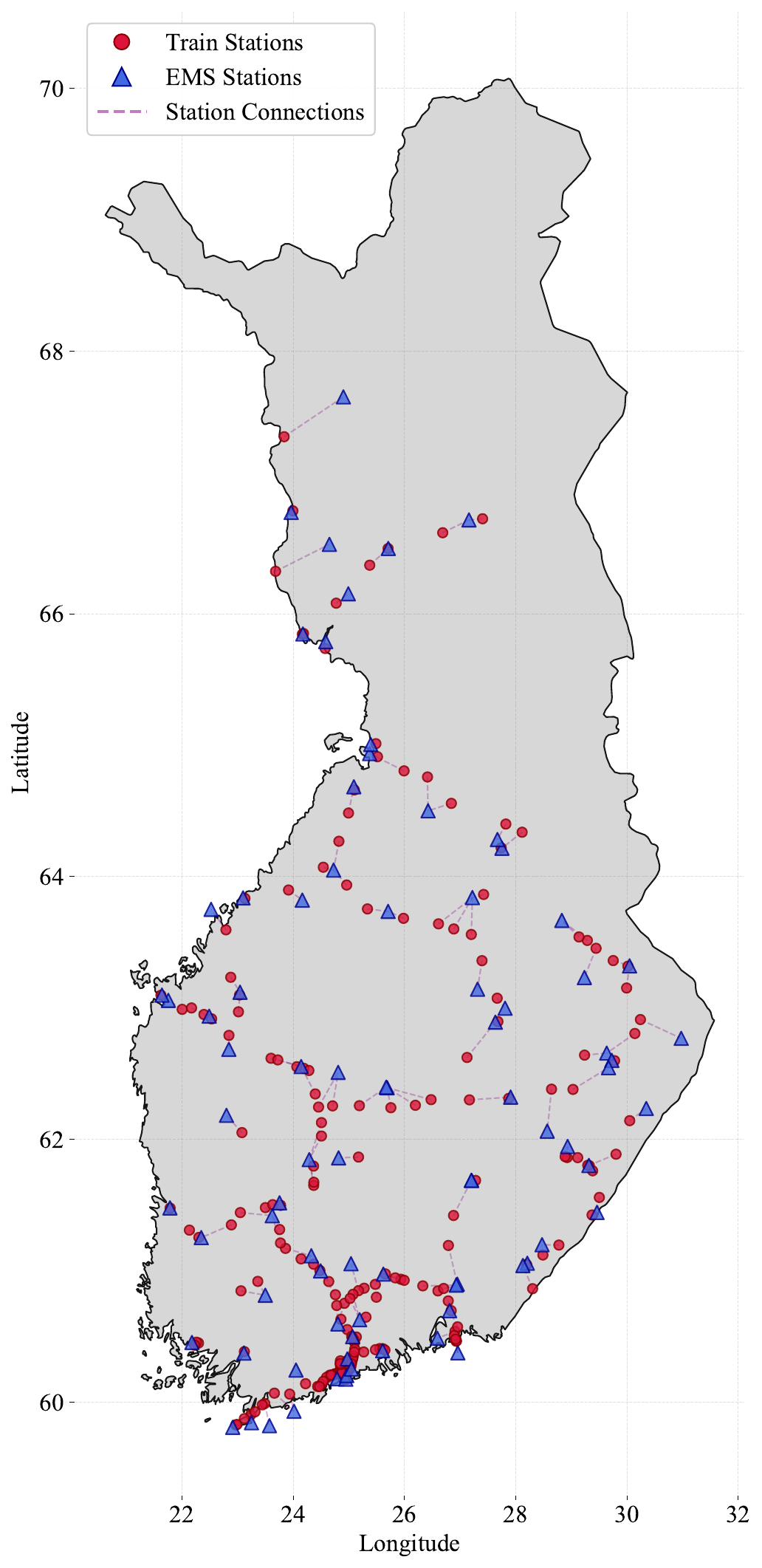}
    \caption{Spatial pairing of train stations with their nearest EMS across the Finnish railway network. Red circles denote train stations, blue triangles denote EMS, and dashed lines connect each train station to the EMS assigned to it by the Haversine nearest-neighbor criterion. Coordinates are given in decimal degrees (WGS84). Base map made with Natural Earth; free vector and raster map data available at \url{naturalearthdata.com}.}
    \label{fig:train_EMS}
    \vspace{-16pt}
\end{figure}

This two-stage approach ensured that each train operation was matched with the most spatially and temporally relevant meteorological conditions, maximizing the reliability of the integrated dataset while preventing data loss from the primary train records. Figure~\ref{fig:train_EMS} shows each train station and the closest EMS where weather data was fetched.

\subsection{Missing Data Weather Mitigation Strategy}
As discussed in Section~\ref{subsec:fmi}, the environmental meteorological infrastructure exhibits heterogeneity in sensor deployment, in which not all EMS are equipped to measure the complete suite of meteorological parameters. To mitigate the impact of missing weather features in the merged dataset, a spatial fallback algorithm was implemented.

The algorithm operates as follows:

\begin{enumerate}
    \item \textit{Feature verification:} For each train record matched to its nearest EMS, the availability of the required meteorological features is verified.
    
    \item \textit{Radial search:} When a specific weather parameter is unavailable at the nearest EMS, the algorithm initiates a radial search for alternative EMS units within a 50 km radius that possess the missing measurement capability.
    
    \item \textit{Data retrieval:} If one or more alternative stations are identified, the weather data from the nearest available station measuring the required parameter is retrieved and incorporated into the dataset.
    
    \item \textit{Missing value designation:} If no stations within the search radius provide the missing measurement, the corresponding feature value is designated as missing in the dataset.
\end{enumerate}

This approach represents a pragmatic balance between spatial representativeness and data completeness. The 50 km radius threshold ensures that retrieved meteorological data remain reasonably representative of conditions at the train section location while maximizing feature availability. The strategy acknowledges that localized weather measurements from a more distant station are preferable to the systematic absence of critical meteorological variables, particularly for regional-scale weather phenomena that exhibit spatial continuity within this distance range.

\subsection{Dropping Unnecessary Columns}

Following dataset merging, a feature selection procedure was performed to remove columns that provided negligible informational content for the predictive task. Specifically, columns that exhibit one or more of the following characteristics were removed: (1) identifier fields with no predictive value, (2) duplicate features that contain identical information, and (3) administrative metadata not related to the underlying phenomena. This reduction in dimensionality improved both computational efficiency and model interpretability by focusing analysis on domain-relevant features.

\subsection{Handling Missing Data}
Missing values were addressed through a hierarchical four-strategy approach:
\begin{enumerate}
    \item \textit{List-wise Deletion:} Applied when (1) timestamps were missing, as temporal integrity is essential for time-series analysis; (2) target features (delay indicators) were missing, since imputing targets introduces training bias (discussed in Section \ref{subsec:final_dataset}); and (3) all weather features were simultaneously missing, as these observations lack predictive context.
    
    \item \textit{Weather Feature-wise Deletion:} Weather columns exceeding 70\% missingness were removed, as systematic sensor failures create sparse features that contribute more noise than signal. Table~\ref{tab:missing_data} shows weather features and their missing data percentages after applying the weather mitigation strategy.
   
    \item \textit{Constant Value Imputation:} Boolean features (\texttt{trainStopping}, \texttt{commercialStop}) with missing values were set to \texttt{False} (encoded as 0), based on domain knowledge that absent flags typically indicate feature negation in transportation datasets.
    
    \item \textit{Month-Specific Median Imputation:} Weather features passing the sparsity threshold were imputed using monthly medians to account for seasonal variability. The median was chosen for its robustness against sensor outliers.
\end{enumerate}

\begin{table}[!hbtp]
\vspace{-0.4cm}
\small
\centering
\caption{Missing data for weather features (2018-2024).}
\label{tab:missing_data}
\begin{tabular}{lrr}
\toprule
\textbf{Feature} & \textbf{Missing Count} & \textbf{Missing (\%)} \\
\midrule
Precipitation amount & $33{,}506{,}086$ & 86.91 \\
Cloud amount & $13{,}048{,}473$ & 33.85 \\
Wind direction & $13{,}000{,}331$ & 33.72 \\
Pressure (msl) & $6{,}976{,}767$ & 18.10 \\
Dew-point temperature & $3{,}219{,}117$ & 8.35 \\
Air temperature & $3{,}182{,}196$ & 8.25 \\
Relative humidity & $3{,}159{,}488$ & 8.20 \\
Snow depth & $2{,}378{,}448$ & 6.17 \\
Precipitation intensity & $2{,}320{,}256$ & 6.02 \\
Horizontal visibility & $2{,}251{,}748$ & 5.84 \\
Wind speed & $2{,}153{,}097$ & 5.59 \\
Gust speed & $2{,}153{,}067$ & 5.58 \\
\midrule
\textbf{Total observations} & \textbf{$38{,}551{,}161$} & \textbf{---} \\
\bottomrule
\end{tabular}
\vspace{-0.4cm}
\end{table}

\subsection{Converting Temporal Features to Cyclical Approach}

Temporal features such as hours, months, and days of the week exhibit inherent cyclical properties that are not adequately captured by their raw numerical representations. When these features are encoded as linear integers, machine learning models encounter the time wraparound problem~\cite{Cai2020}, where values at cycle boundaries are treated as maximally distant despite their temporal proximity. For instance, 23:00 (11 PM) and 00:00 (midnight) are separated by only one hour temporally, yet their numerical representation suggests a distance of 23 units, potentially misleading gradient-based and distance-based learning algorithms.

The temporal values of hour, month and day\_of\_week are transformed into $(\text{hour\_sin}, \text{hour\_cos})$, $(\text{month\_sin}, \text{month\_cos})$, and $(\text{day\_week\_sin}, \text{day\_week\_cos})$, respectively. These encodings effectively communicate to the model that these values reside at the terminal positions of their respective cycles, facilitating accurate learning of wraparound patterns.

Three distinct transformation stages were implemented to encode temporal features 
using sine-cosine encoding at different granularities. We employed the standard 
cyclical encoding formula:
\begin{equation}
\text{encoded\_feature} = \sin\left(\frac{2\pi t}{P}\right), \quad 
\cos\left(\frac{2\pi t}{P}\right),
\end{equation}
where $t$ represents the temporal value and $P$ is the corresponding period. 
Specifically, the hour feature with $t \in \{0, 1, \ldots, 23\}$ was encoded with 
$P=24$; the month feature with $t \in \{1, 2, \ldots, 12\}$ with $P=12$; and the 
day-of-week feature with $t \in \{1, 2, \ldots, 7\}$ (where 1 = Monday) with $P=7$. 
The original hour column was subsequently removed, as the sine-cosine representation 
suffices for all model architectures. In contrast, the original month and day-of-week 
columns were retained alongside their encoded variants to support diverse model 
requirements.

\subsection{Additional Data Integrity Steps}\label{subsec:additional_steps}

\subsubsection{Removing Duplicates}

Duplicate records were identified and removed to ensure that the dataset did not contain redundant observations. Duplicates were detected across all feature dimensions using an exact matching approach. Duplicates represented 23.24\% of the dataset and were completely removed.

\subsubsection{Splitting Data}
The dataset was partitioned into training and test sets using an 80/20 split, allocating 80\% of observations for model training and 20\% for evaluation. Crucially, this step was performed before any scaling operations to prevent data leakage. This division ensures that model performance is assessed on previously unseen data, providing an unbiased estimate of generalization capability.

\subsubsection{Scaling Weather Data}

Applied only after splitting data to prevent data leakage, scaling parameters were computed solely from the training partition only and subsequently applied to both training and test sets. This ensures that no statistical information from the test set influences the transformation.

To ensure comparability between meteorological variables with disparate measurement scales and units, a robust scaling transformation was applied to the weather features. Unlike standard normalization, which employs mean and standard deviation, this study utilized RobustScaler from the scikit-learn library. This approach is advantageous for weather data because meteorological measurements are susceptible to outliers and extreme values that can distort conventional scaling methods. By leveraging quartile-based measures, RobustScaler minimizes the influence of such anomalies, ensuring that scaling remains representative of the typical data distribution. 

The RobustScaler normalization can be defined as
\begin{equation}
x_{\text{scaled}} = \frac{x - Q_1(X_{\text{train}})}{Q_3(X_{\text{train}}) - Q_1(X_{\text{train}})},
\end{equation}
where $Q_1$ denotes the first quartile (25th percentile) and $Q_3$ denotes the third quartile (75th percentile), both calculated from the training partition. Quartiles are obtained by sorting the training observations and identifying the values that divide the distribution into four equal parts, such that 25\% of observations fall below $Q_1$ and 75\% fall below $Q_3$. The denominator— of the equation, the interquartile range, represents the range spanning the central 50\% of the training data, a statistic fundamentally resistant to extreme values.

\section{Proposed Final Dataset} \label{subsec:final_dataset}

Table \ref{tab:final_dataset_descriptions} presents a comprehensive overview of the feature set comprising our final dataset for the train delay prediction in Finland. The dataset encompasses 138 distinct features organized into three primary categories: target features, operational features, and weather features. The weather features are further divided into eleven base measurements and two families of engineered variables: 69 rolling-window aggregations (Table~\ref{tab:rolling_windows}) and 40 weather scenario indicators (Table~\ref{tab:weather_scenarios}). Approximately 86.2\% of these features are derived columns, generated through feature engineering techniques to enhance the predictive capability of our models.

\begin{table*}[!hbtp]
\centering
\small
\caption{Final Dataset Feature Descriptors}
\label{tab:final_dataset_descriptions}
\begin{tabular}{cl>{\raggedright\arraybackslash}p{9.2cm}c}
\toprule
\textbf{No.} & \textbf{Feature Name} & \textbf{Description} & \textbf{Derived?} \\
\midrule
\multicolumn{4}{l}{\textbf{Target Features}} \\
\midrule
1 & differenceInMinutes & Delay in minutes (actual - scheduled) & No \\
2 & differenceInMinutes\_offset & Delay in minutes (delay offset removed from 1st station only) & Yes \\
3 & differenceInMinutes\_eachStation\_offset & Delay in minutes (delay offset removed from all stations) & Yes \\
4 & trainDelayed & Boolean indicating if train is delayed & Yes \\
5 & cancelled & Boolean indicating if train was cancelled & No \\
\midrule
\multicolumn{4}{l}{\textbf{Operational Features}} \\
\midrule
6 & trainStopping & Boolean indicating if train stopped at station & No \\
7 & commercialStop & Boolean indicating if stop is commercial & No \\
8 & hour & Hour of the day (0-23) & No \\
9 & hour\_sin & Sine component of cyclical hour encoding & Yes \\
10 & hour\_cos & Cosine component of cyclical hour encoding & Yes \\
11 & month & Month of the year (1-12) & No \\
12 & month\_sin & Sine component of cyclical month encoding & Yes \\
13 & month\_cos & Cosine component of cyclical month encoding & Yes \\
14 & day\_of\_week & Day of the week (1-7) & No \\
15 & day\_week\_sin & Sine component of cyclical day of week encoding & Yes \\
16 & day\_week\_cos & Cosine component of cyclical day of week encoding & Yes \\
17 & day\_of\_month & Day of the month (1-31) & No \\
18 & causes\_related\_to\_weather & Boolean indicating if the recorded delay cause is weather-related & Yes \\
\midrule
\multicolumn{4}{l}{\textbf{Weather Features (base)}} \\
\midrule
19 & Air temperature & Air temperature (°C) & No \\
20 & Wind speed & Wind speed (m/s) & No \\
21 & Gust speed & Wind gust speed (m/s) & No \\
22 & Wind direction & Wind direction (degrees) & No \\
23 & Relative humidity & Relative humidity (\%) & No \\
24 & Dew-point temperature & Dew-point temperature (°C) & No \\
25 & Precipitation intensity & Precipitation intensity (mm/h) & No \\
26 & Snow depth & Snow depth (cm) & No \\
27 & Pressure (msl) & Mean sea level pressure (hPa) & No \\
28 & Horizontal visibility & Horizontal visibility (m) & No \\
29 & Cloud amount & Cloud amount (oktas, 0-9) & No \\
\midrule
\multicolumn{4}{l}{\textbf{Weather Features (engineered)}} \\
\midrule
30--69 & weather\_scenario\_$\ast$ & 10 adverse-weather scenario indicators at 4 temporal windows (Table~\ref{tab:weather_scenarios}) & Yes \\
70--138 & \textit{rolling aggregations} & Rolling-window statistics of weather variables over 12/24/72\,h (Table~\ref{tab:rolling_windows}) & Yes \\
\bottomrule
\end{tabular}
\end{table*}

The \textbf{operational features} capture the temporal patterns and characteristics of the train service. Two boolean variables, \textit{trainStopping} and \textit{commercialStop}, distinguish stopping behaviors. Recognizing the cyclical nature of temporal data, we applied trigonometric encoding to month, hour, and day of week, generating sine and cosine components that preserve circular relationships. Importantly, the primary temporal columns (\textit{hour}, \textit{month}, \textit{day\_of\_week}, \textit{day\_of\_month}) are retained in the final dataset for generic use purposes, as tree-based algorithms may better utilize these raw features, while cyclical encodings are more suitable for numeric/regression-based techniques. An additional boolean feature, \textit{causes\_related\_to\_weather}, is derived from the Digitraffic \textit{causes} array and flags whether the officially recorded cause of a delay is weather-related, providing a label for weather-impact studies.

The \textbf{weather features} comprise eleven \textit{base} meteorological measurements from the FMI (Features 19--29, Table~\ref{tab:final_dataset_descriptions}), reporting the instantaneous conditions at each train section, together with the two families of engineered weather features introduced in Section~\ref{subsubsec:derived_weather}: 69 rolling-window aggregations (Table~\ref{tab:rolling_windows}) and 40 weather scenario indicators (Table~\ref{tab:weather_scenarios}). Collectively, these features enable our models to account for weather-related disruptions, which may constitute a factor in railway delays.

The \textbf{target features} constitute the dependent variables and delay-related indicators that our models aim to predict or utilize for classification tasks:

\begin{itemize}
    \item \textit{differenceInMinutes}: The primary target variable representing the raw delay calculated as the difference between actual and scheduled arrival times.
    
    \item \textit{differenceInMinutes\_offset}: An offset feature representing the delay in minutes excluding the initial delay accumulated at the first station, thus isolating delays that develop during the journey.
    
    \item \textit{differenceInMinutes\_eachStation\_offset}: An offset feature that isolates the delay contribution specific to each individual station by removing propagated delays from previous stops.
\end{itemize}

\begin{figure}[!hbtp]
    \centering
\includegraphics[width=0.98\columnwidth]{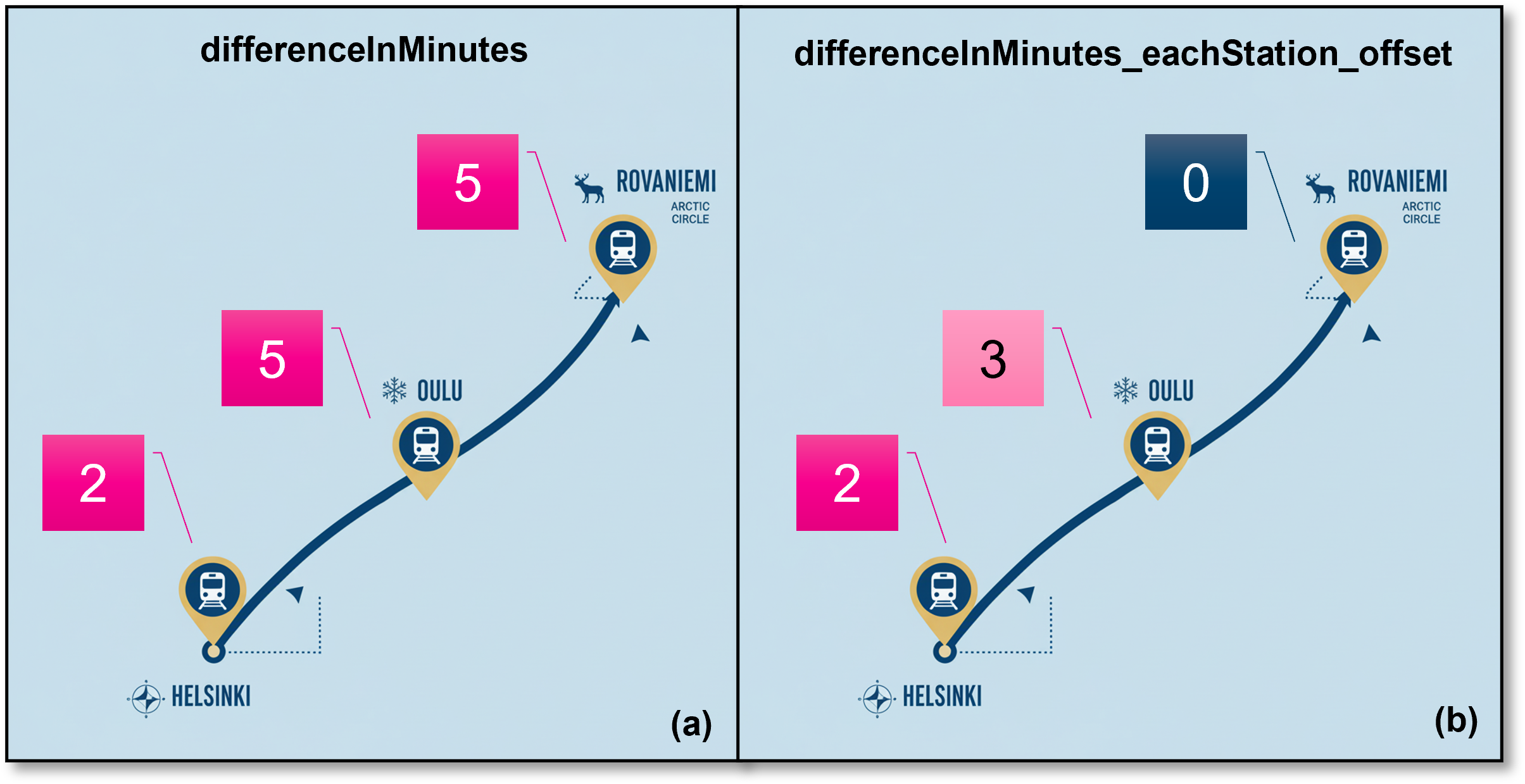}
    \caption{Delay target variables comparison: (a) \textit{differenceInMinutes}, (b) \textit{differenceInMinutes\_eachStation\_offset}}
    \label{fig:delay_explained}
    \vspace{-16pt}
\end{figure}

To clarify the construction of \textit{differenceInMinutes\_eachStation\_offset}, consider the example illustrated in Figure \ref{fig:delay_explained}. This figure depicts a hypothetical train route from Helsinki to Rovaniemi via Oulu. The number above each station represents the delay in minutes at that station. 
Figure \ref{fig:delay_explained}a shows that the train departed Helsinki 2 minutes late, arrived at Oulu 5 minutes late, and reached Rovaniemi 5 minutes late. However, these values represent the total accumulated delay at each point, not the delay introduced on each segment.
Figure \ref{fig:delay_explained}b illustrates the delay offset for each station segment after removing the inherited delay from previous stations. If the train departed Helsinki 2 minutes late and arrived at Oulu 5 minutes late, the delay introduced specifically in the Helsinki–Oulu segment is 3 minutes (5 - 2 = 3), not 5 minutes. Similarly, since the train arrived at both Oulu and Rovaniemi with the same 5-minute delay, the Oulu–Rovaniemi segment did not introduce additional delay (5 - 5 = 0). The \textit{differenceInMinutes\_eachStation\_offset} target feature thus represents the incremental delay added to each route segment, isolating the delay contribution between consecutive stations rather than the cumulative delay from the origin.

Additionally, two boolean indicators facilitate binary classification tasks and enable analysis of severe disruptions in rail operations:
\begin{itemize}
    \item \textit{trainDelayed}: A boolean feature containing True/False information using a 5-minute threshold to classify whether a train is delayed.
    
    \item \textit{cancelled}: A boolean feature that contains raw information that indicates whether a train was canceled by the operator.
\end{itemize}

\subsection{Dataset Usage and Configuration}
The dataset has been designed with maximum flexibility in mind, allowing researchers and practitioners to adapt it to various predictive modeling scenarios. Users have complete autonomy to select relevant weather features according to their specific research questions and to choose appropriate target variables based on their analytical objectives. This modular design enables the dataset to serve multiple use cases, from weather-impact studies to operational delay prediction systems.

A correlation analysis was conducted to identify potential feature redundancies and inform feature selection strategies. The analysis revealed several strong correlations between weather variables: gust speed and wind speed exhibit very high correlation ($0.946$), dew-point temperature shows strong correlations with air temperature ($0.898$) and relative humidity ($0.595$), while snow depth demonstrates moderate correlations with dew-point temperature ($0.575$) and horizontal visibility ($0.571$). The heatmap enables identification of strongly correlated features, detection of multicollinearity issues, and insights into which weather conditions most significantly impact train operations. These highly correlated features could be candidates for removal during model training to reduce multicollinearity and improve model efficiency, though they remain in the dataset to allow flexibility in feature selection strategies for different modeling approaches.


The dataset includes both numeric and binary target variables to accommodate different modeling approaches and research objectives. Numeric targets include delay measurements (Features 1--3, Table~\ref{tab:final_dataset_descriptions}), while binary targets comprise operational status indicators (Features 4--5, Table~\ref{tab:final_dataset_descriptions}).

\begin{figure}[!hbtp]
    \centering
    \begin{subfigure}[b]{\columnwidth}
        \centering
        \includegraphics[width=\columnwidth]{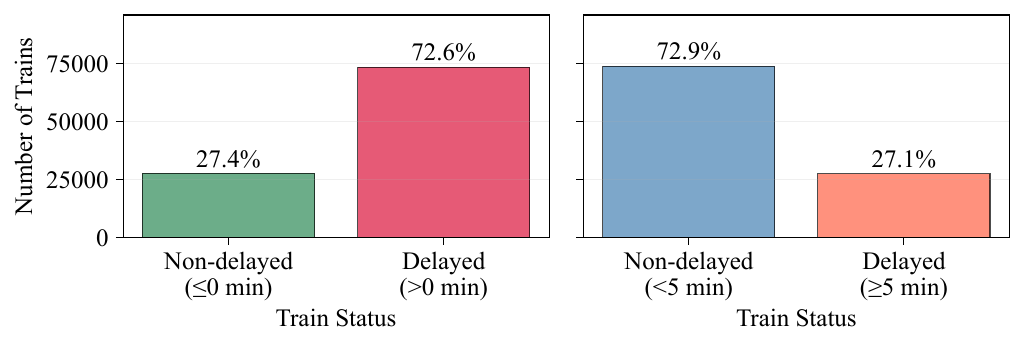}
        \caption{Delay distribution with propagation delay (\textit{differenceInMinutes} target variable).}
        \label{fig:delay_distribution}
    \end{subfigure}
    
    \begin{subfigure}[b]{\columnwidth}
        \centering
        \includegraphics[width=\columnwidth]{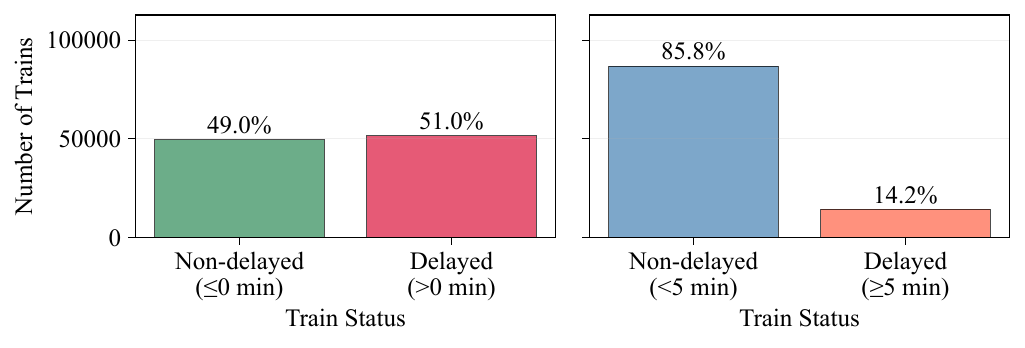}
        \caption{Delay distribution per station (no delay propagation).}
        \label{fig:delay_distribution_per_station}
    \end{subfigure}
    \caption{Target feature (delay in minutes) distributions (\textit{differenceInMinutes\_eachStation\_offset} target variable).}
    \label{fig:target_distributions}
    \vspace{-16pt}
\end{figure}

Figure~\ref{fig:target_distributions} compares delay distributions under two different formulations for Oulu asema (Oulu station), considering all long-distance trains that pass through Oulu during the seven-year period. Figure~\ref{fig:delay_distribution} shows \texttt{differenceInMinutes}, which measures cumulative delay from the origin. Here, 72.6\% of observations show some delay (zero threshold), and 27.1\% exceed five minutes. Figure~\ref{fig:delay_distribution_per_station} shows \texttt{differenceInMinutes\_eachStation\_offset}, which isolates the delay generated at each station by removing propagation effects. This formulation yields 51\% delayed observations (zero threshold) and 14.2\% exceeding five minutes.

The difference reflects how delays accumulate: most observed delays result from propagation rather than station-specific causes. The choice of target variable depends on the modeling objective, as \texttt{differenceInMinutes} suits end-to-end delay prediction for passenger information systems, and \texttt{differenceInMinutes\_eachStation\_offset} identifies local operational issues and weather impacts at specific stations. The class imbalance also differs significantly between formulations, requiring different handling strategies for classification tasks.


\section{Preliminary Baseline Experiment}

To illustrate the dataset's utility for ML applications, we conducted a preliminary baseline prediction experiment using XGBoost regression, a gradient boosting algorithm widely employed for predictive modeling tasks across diverse domains~\cite{chen2016xgboost}. This experiment is intended as an illustrative demonstration of how the dataset can be used, rather than an attempt to obtain an optimized or state-of-the-art delay-prediction model.

\subsection{Experimental Setup}

The experiment focused on long-distance trains passing through Oulu asema during the 2018--2024 period, comprising $101{,}146$ observations. The target variable was \texttt{differenceInMinutes\_eachStation\_offset}, which isolates station-specific delay contributions by removing propagated delays from previous stops. This formulation enables the model to learn local delay patterns influenced by weather conditions rather than cumulative delay propagation.

The feature set comprised 18 variables spanning operational and meteorological domains: \texttt{trainStopping}, cyclical temporal encodings (\texttt{month\_sin}, \texttt{month\_cos}, \texttt{hour\_sin}, \texttt{hour\_cos}, \texttt{day\_of\_week}, \texttt{day\_week\_sin}, \texttt{day\_week\_cos}), and ten weather features (air temperature, wind speed, gust speed, wind direction, relative humidity, dew-point temperature, precipitation intensity, snow depth, pressure, and horizontal visibility).

Hyperparameter optimization was performed using randomized search with 50 iterations and 5-fold cross-validation on the training partition. The dataset was split 80/20 for training and testing, with weather features scaled using RobustScaler parameters derived exclusively from the training set, as detailed in Section~\ref{subsec:additional_steps}.

\subsection{Result and Discussion}
\begin{figure}[!hbtp]
    \centering
    \includegraphics[width=0.98\columnwidth]{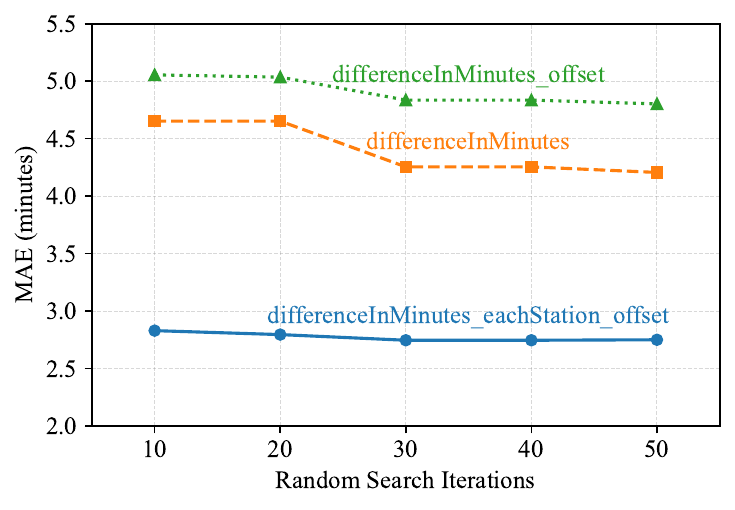}
    \caption{MAE comparison of XGBoost models across random search iterations for three delay prediction targets.}
    \label{fig:mae_comparison}
    \vspace{-16pt}
\end{figure}

Figure~\ref{fig:mae_comparison} presents the Mean Absolute Error (MAE) performance of XGBoost models across three delay prediction targets over 50 random search iterations for hyperparameter optimization. 

The station-specific delay target (\texttt{differenceInMinutes\_eachStation\_offset}) demonstrates superior predictive performance, achieving an MAE of 2.73 minutes on the test set for Oulu asema. This represents a substantial improvement over cumulative delay prediction targets: \texttt{differenceInMinutes} (4.21 minutes MAE) and \texttt{differenceInMinutes\_offset} (4.81 minutes MAE). The performance gap illustrates that station-specific delays, which isolate local operational conditions, are inherently more predictable than accumulated delays that propagate through the network. All models exhibit convergence by 30 iterations, with minimal performance variation thereafter.

Direct comparison with existing literature is challenging due to differences in railway systems, prediction targets, and evaluation contexts. Huang et al.~\cite{Huang2020} employed different algorithms on Chinese high-speed railways; for long-distance trains, Random Forest and Support Vector Machine achieved MAE values of 2.16 minutes, while their proposed FCLL-Net deep learning architecture achieved 1.87 minutes (one-step predictor). Li et al.~\cite{Li2021} applied Random Forest and XGBoost to Dutch conventional railways, achieving MAE values of 1.71 and 1.72 minutes for delays exceeding 3 minutes. However, both the Chinese high-speed and Dutch conventional systems operate predominantly on double-track infrastructure with milder climatic conditions. The Finnish context introduces unique challenges, including extreme seasonal weather variations (temperatures ranging from $-40$°C to $+30$°C) and a predominantly single-track network with different operational characteristics.

As a preliminary demonstration, this experiment is deliberately limited in scope. It considers a single station (Oulu asema) with a single learning algorithm, reports a single error metric, and uses only the base weather and operational features. The engineered weather-scenario indicators and rolling-window aggregations, a systematic assessment of the incremental predictive value of weather (e.g., through with- and without-weather ablations), and a network-wide evaluation across multiple stations are all left for future work. The intent here is to establish that FI-TW supports end-to-end machine-learning workflows and to provide a reproducible reference point, rather than to deliver an optimized delay-prediction model.

\section{Conclusions} \label{sec:conclusions}

This paper presented FI-TW, the Finland Integrated Train--Weather dataset, which is, to the best of our knowledge, the first publicly available resource integrating Finnish railway operational data with synchronized meteorological observations from 2018-2024. The dataset combines train schedules and delay records from Digitraffic with weather measurements from 209 FMI stations, covering approximately 38.5 million observations across Finland's $5{,}915$-kilometer rail network.
Exploratory analysis revealed distinct seasonal patterns, with winter months exhibiting delay rates exceeding 25\% compared to below 20\% in summer, and geographic clustering of high-delay corridors in central and northern Finland.
The dataset offers multiple target variable formulations, cumulative delays for passenger information systems and station-specific offset delays for isolating local operational impacts, enabling diverse modeling approaches from binary classification to regression-based prediction. A baseline XGBoost experiment achieved MAE of 2.73 minutes for station-specific delay prediction, demonstrating the dataset's utility for machine learning applications.

Several research directions emerge from this work. The temporal structure of train journeys, where delays propagate through successive stations, suggests that sequence modeling architectures such as Long Short-Term Memory networks and Transformers may capture delay dynamics more effectively than traditional approaches. Graph Neural Networks could exploit network-wide dependencies between stations and routes. Extending the dataset with streaming data pipelines and weather forecasts would enable real-time prediction systems capable of advance warning rather than post-hoc analysis.
Finally, causal inference methods could quantify the direct impact of specific meteorological conditions on railway performance, while incorporating additional data sources such as passenger flows, maintenance records, or rolling stock information could further enhance predictive accuracy.

\section{Data Availability} \label{sec:data_availability}
The complete dataset is publicly available on Kaggle\footnote{\href{https://www.kaggle.com/datasets/viniborin/finland-integrated-train-weather-dataset-fi-tw}{https://doi.org/10.34740/kaggle/dsv/14124620}}. The data spans from January 2018 to December 2024 and is organized in Apache Parquet files separated by month and year, facilitating selective access and incremental processing. Supporting source code repositories for data collection, visualization, and preprocessing are listed on the Kaggle page.

\bstctlcite{IEEEexample:BSTcontrol}
\bibliographystyle{IEEEtran}
\bibliography{IEEEabrv,references}

\end{document}